\documentclass{article}
\usepackage{hello}

\usepackage[margin=1in]{geometry}

\usepackage{graphicx}
\usepackage{booktabs}
\usepackage{array}
\usepackage{amsmath}
\usepackage{amssymb}
\usepackage{amsfonts}
\usepackage{multirow}
\usepackage{verbatim}
\usepackage{caption}
\usepackage{longtable}
\usepackage{supertabular}
\usepackage{float}

\usepackage{enumitem}
\usepackage{tablefootnote}
\usepackage[round,semicolon]{natbib}
\usepackage{xcolor}
\usepackage{colortbl}
\usepackage{xspace}
\usepackage{textcomp}
\usepackage{makecell}
\usepackage{multirow}
\usepackage{lscape} 
\usepackage{siunitx}

\usepackage{amssymb}% http://ctan.org/pkg/amssymb
\usepackage{pifont}% http://ctan.org/pkg/pifont
\usepackage{scrextend}

\usepackage{array}
\usepackage{tgpagella}
\usepackage{latexsym}
\usepackage[T1]{fontenc}
\usepackage[utf8]{inputenc}
\usepackage{microtype}
\definecolor{mydarkblue}{rgb}{0,0.08,0.45}
\usepackage[colorlinks,citecolor=mydarkblue,urlcolor=mydarkblue,linkcolor=mydarkblue]{hyperref}
\usepackage{url}            % simple URL typesetting
\usepackage{nicefrac}       % compact symbols for 1/2, etc.
\usepackage{changepage}
\usepackage{xargs}          % Use more than one optional parameter in a new commands
\usepackage{wrapfig,lipsum,booktabs}
\usepackage{longtable}
\usepackage{subcaption}
\usepackage{endnotes}

\usepackage{pgfplots}
\usetikzlibrary{pgfplots.groupplots}
\pgfplotsset{compat=1.3}
\usepackage{tikz}
\usetikzlibrary{patterns}

\usepackage[most]{tcolorbox}

\usepackage[capitalize,noabbrev]{cleveref}
\crefname{section}{Section}{\S\S}
\Crefname{section}{Section}{\S\S}
\crefname{table}{Table}{Tables}
\crefname{figure}{Figure}{Figures}
\crefname{algorithm}{Algorithm}{}
\crefname{equation}{eq.}{}
\crefname{appendix}{Appendix}{}
\crefformat{section}{Section #2#1#3}
\usepackage{multicol}
\usepackage{multirow}

\usepackage{tcolorbox}
\usepackage{titlesec}
\titleformat*{\section}{\large\bfseries}

\definecolor{battleshipgrey}{rgb}{0.3, 0.3, 0.3}
\definecolor{brilliantrose}{rgb}{1.0, 0.33, 0.64}
\definecolor{americanrose}{rgb}{1.0, 0.01, 0.24}
\definecolor{jweigreen}{rgb}{0,0.45,0.24}
\definecolor{bluegray}{rgb}{0.1, 0.1, 0.4}
\definecolor{ao(english)}{rgb}{0.0, 0.5, 0.0}
\definecolor{blanchedalmond}{rgb}{1.0, 0.92, 0.8}
\definecolor{atomictangerine}{rgb}{1.0, 0.6, 0.4}
\definecolor{chocolate(web)}{rgb}{0.82, 0.41, 0.12}
\definecolor{bananayellow}{rgb}{1.0, 0.88, 0.21}
\definecolor{goldenbrown}{rgb}{0.6, 0.4, 0.08}
\definecolor{aliceblue}{rgb}{0.94, 0.97, 1.0}
\definecolor{beige}{rgb}{0.96, 0.96, 0.86}
\definecolor{babyblue}{rgb}{0.54, 0.81, 0.94}
\definecolor{camel}{rgb}{0.76, 0.6, 0.42}
\definecolor{cinnamon}{rgb}{0.82, 0.41, 0.12}
\definecolor{deepskyblue}{rgb}{0.0, 0.75, 1.0}
\definecolor{frenchblue}{rgb}{0.0, 0.45, 0.73}
\definecolor{classicrose}{rgb}{0.98, 0.8, 0.91}
\definecolor{frenchrose}{rgb}{0.96, 0.29, 0.54}
\definecolor{frenchlilac}{rgb}{0.53, 0.38, 0.56}
\definecolor{frenchbeige}{rgb}{0.65, 0.48, 0.36}
\definecolor{verylightgreen}{RGB}{240, 255, 235}
\definecolor{verylightred}{RGB}{255, 235, 235}
\definecolor{verylightyellow}{RGB}{255, 254, 235}
\definecolor{dt}{gray}{0.7}

\definecolor{forestgreen}{HTML}{2e7d43}
\definecolor{color1}{HTML}{FF9999}
\definecolor{color2}{HTML}{FF6666}
\definecolor{color3}{HTML}{FF3333}
\definecolor{color4}{HTML}{E60000}
\definecolor{color5}{HTML}{B30000}
\definecolor{color6}{HTML}{8CD98C}
\definecolor{color7}{HTML}{53c653}
\definecolor{color8}{HTML}{39ac39}
\definecolor{color9}{HTML}{2d862d}
\definecolor{color10}{HTML}{206020}
\definecolor{color11}{HTML}{cca300}

\usepackage{minitoc}

\usepackage{graphicx}
\usepackage{tabularx}

\title{\textbf{TouchStone: Evaluating Vision-Language Models by Language Models}
}

\author{
$^1$Shuai Bai \hspace{6mm} $^{1,2}$Shusheng Yang \hspace{6mm} $^{1}$Jinze Bai  \hspace{6mm} $^{1}$Peng Wang\hspace{6mm} $^{1,3}$Xingxuan Zhang \\
\large{}
 $^{1}$Junyang Lin \hspace{6mm} $^{2}$Xinggang Wang \hspace{6mm} $^{1}$Chang Zhou$^{\dag}$ \hspace{6mm}  $^{1}$Jingren Zhou
\\
 $^{1}$Alibaba Group,  $^{2}$Huazhong University of Science and Technology,  $^{3}$Tsinghua University
}

\begin{document}
\maketitle

\begin{abstract}
Large vision-language models (LVLMs) have recently witnessed rapid advancements, exhibiting a remarkable capacity for perceiving, understanding, and processing visual information by connecting visual receptor with large language models (LLMs).
However, current assessments mainly focus on recognizing and reasoning abilities, lacking direct evaluation of conversational skills and neglecting visual storytelling abilities.
In this paper, we propose an evaluation method that uses strong LLMs as judges to comprehensively evaluate the various abilities of LVLMs.
Firstly, we construct a comprehensive visual dialogue dataset TouchStone, consisting of open-world images and questions, covering five major categories of abilities and 27 subtasks. This dataset not only covers fundamental recognition and comprehension but also extends to literary creation.
Secondly, by integrating detailed image annotations we effectively transform the multimodal input content into a form understandable by LLMs.
This enables us to employ advanced LLMs for directly evaluating the quality of the multimodal dialogue without requiring human intervention.
Through validation, we demonstrate that powerful LVLMs, such as GPT-4, can effectively score dialogue quality by leveraging their textual capabilities alone, aligning with human preferences.
We hope our work can serve as a touchstone for LVLMs' evaluation and pave the way for building stronger LVLMs.
The evaluation code is available at \url{https://github.com/OFA-Sys/TouchStone}.
\end{abstract}

{\let\thefootnote\relax\footnotetext{ $^\dag$Corresponding author}}

\section{Introduction}
The utilization of large language models (LLMs)~\citep{zhang2022opt,gao2023llama,gpt3,gpt4,anil2023palm} in the domain of chatbots~\citep{ouyang2022intructgpt,vicuna2023} has exhibited remarkable prowess in various aspects such as language comprehension, generation, and interaction.
The extension of GPT-4~\citep{gpt4} to encompass LLMs has further facilitated the rapid development of large vision-language models (LVLMs).
Recently, several LVLMs~\citep{dai2023instructblip,li2023otter,zhu2023minigpt,su2023pandagpt,li2023blip,liu2023visual,ye2023mplug,gao2023llamaadapterv2} have been proposed with the objective of extending the capabilities of pure-text chatbots to incorporate multimodal chatbots.
This is achieved through the alignment of visual encoders with LLMs and the application of visual instruction tuning techniques.
However, it is noteworthy that the evaluation of these recent LVLMs has predominantly focused on the human evaluation of generation quality within a limited subset of questions, thus lacking a comprehensive quantitative evaluation.
\begin{figure*}[!ht]
\includegraphics[width= 16cm]{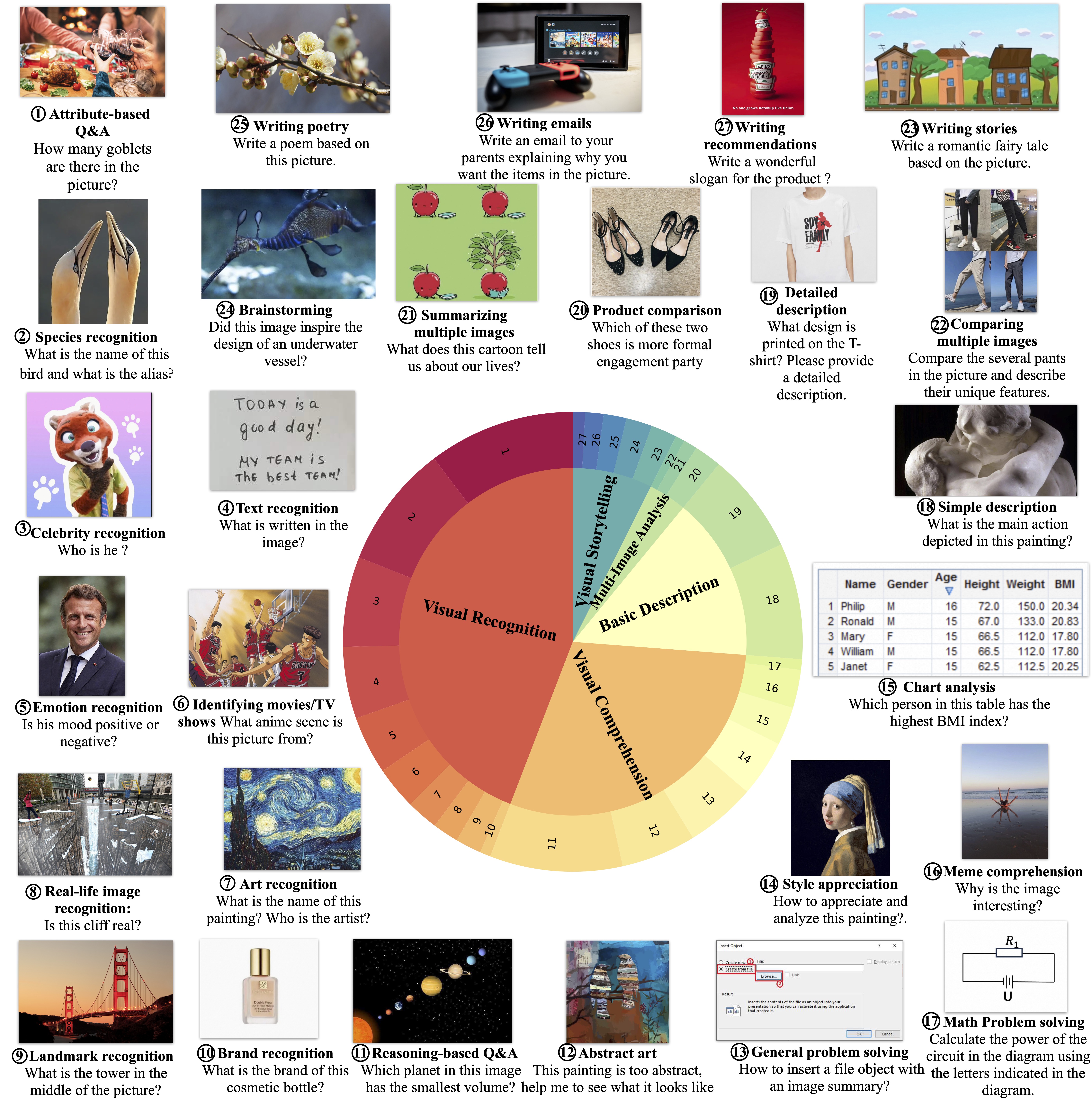}
   \caption{Overview of the dataset distribution and some examples. TouchStone encompasses five major categories and 27 subcategories of questions originating from open scenes, covering the spectrum from recognition and description to understanding and generation.}
\label{dataset}
\end{figure*}

Recent developments in LLM evaluation methodologies~\citep{zheng2023judging}, utilizing automated model assessment, have shown encouraging potential in terms of efficiency and cost-effectiveness when compared to manual evaluation.
Nevertheless, despite these significant advancements in text-based capabilities, the incorporation of multimodal inputs into LLMs remains constrained and underexplored.

Currently, the evaluation methods for LVLMs primarily involve comparing different models based on a small set of questions or assessing their performance on traditional multimodal tasks such as VQA~\citep{goyal2017making,sidorov2020textcaps}, image captioning~\citep{agrawal2019nocaps,chen2015microsoft}, and image classification~\citep{deng2009imagenet}.
However, traditional task metrics and annotations often have specific stylistic preferences for the sake of evaluation and comparison. These stylistic preferences~\citep{agrawal2019nocaps,chen2015microsoft,goyal2017making} do not necessarily align with human preferences, like VQA and image caption. 
Besides, obtaining human annotator ratings or comparisons for different models' outputs is prohibitive and difficult to further scale up.
Additionally, hallucination lies a crucial obstacle for current LLMs or LVLMs broader application, but how to evaluate LVLMs' hallucination degree is always overlooked in most current LVLM evaluations and remains to be explored.
Therefore, there is an urgent need for automated evaluation techniques that can provide objective and efficient assessments of LVLM performance in open-ended dialogues.
Recently, MME~\citep{fu2023mme} has been proposed to transform questions into binary judgment statements for large model evaluation.
MMBench~\citep{liu2023mmbench} evaluates models based on their accuracy in choosing answers.
However, the binary judgment and the ability to choose an answer may not fully capture the complexity in open-ended real-world dialogues, thereby limiting their suitability as a comprehensive evaluation method.

To tackle these challenges, we propose an automated evaluation method termed \textbf{TouchStone}, which provides a comprehensive assessment of the capabilities of multimodal language models.
The principles of our design are two-fold:

Firstly, in order to evaluate the overall abilities of the models, we construct a comprehensive visual dialogue dataset, encompassing five major categories of abilities and 27 subtasks.
These categories include basic descriptive ability, visual recognition ability, visual comprehension ability, visual storytelling ability, and multi-image analysis ability.
This not only tests the model's recognition and comprehension abilities but also tests its literary creation and analysis abilities.
The images and questions in our dataset are curated in an open-world setting and have been manually annotated and verified.

Secondly, TouchStone involves converting information from other modalities, such as images, into textual forms by utilizing detailed image annotations and descriptions.
This enables the employment of advanced LLMs to directly assess the quality of dialogues without the requirements for human evaluators or visual-augmented LLMs.
To reflect the model's performance in real-world scenarios, we conducted a direct evaluation of the quality of dialogue by comparing its correctness, relevance, and usefulness.
For scoring, we utilize a leading language model as a judge, comparing the responses of different LVLM with the answers generated by GPT-4 using pairwise comparisons.
The response generated by GPT-4 is obtained through the input of fine-grained image annotations and question, and are referred to as GPT4-HA (Human Assisted).
To address positional bias, we incorporate position balancing into our scoring mechanism. Through the comparison with human evaluations, we find that powerful LLMs like GPT-4~\citep{gpt4} can effectively score dialogue quality based solely on their text-based capabilities, while also being able to discern hallucination issues.

Our contributions can be summarized as follows:
\begin{itemize}
\item We curate a diverse visual dialogue dataset that covers five categories of abilities and 27 subtasks, encompassing not only basic recognition and comprehension but also extending to literary creation.
\item TouchStone converts information from other modalities, such as images, into textual forms using detailed annotations. This allows advanced language models to directly assess dialogue quality without manual intervention.
\item We show that GPT-4 can serve as a reasonable evaluator to evaluate the quality of LVLMs' response.
Specifically, in our experiment, we find that GPT-4 shows highly-consistent judgment compared to human preference.

\end{itemize}

\section{Related Work}

\subsection{Large Language Models}
Language pretrained models such as GPT~\citep{gpt2,gpt3}, BERT~\citep{devlin2018bert}, and T5~\citep{raffel2020t5} have demonstrated exceptional performance in a multitude of natural language processing (NLP) tasks, thanks to their extensive pre-training on copious amounts of data. Notably, the GPT-3~\citep{gpt3} model, with its decoder-only architecture, has exhibited impressive zero-shot capabilities as the model size and training data have increased.
Furthermore, the field has witnessed the emergence of increasingly sophisticated large-scale models like OPT~\citep{zhang2022opt}, LLaMA~\citep{touvron2023llama}, and PaLM~\citep{anil2023palm}, which have been meticulously constructed to tackle complex NLP challenges. InstructGPT~\citep{ouyang2022intructgpt} incorporates instruction fine-tuning and reinforcement learning, enabling them to align with human preferences and effectively execute specified instructions.
Moreover, ChatGPT surpasses its predecessors by engaging in conversation interactions and successfully carrying out diverse user commands. It demonstrates the potential to address a wide range of real-world tasks and requirements. 

\subsection{Vision-Language Models}

Extensive research has been undertaken on cross-modal Vision-Language Models (VLMs), utilizing different pre-training tasks like mask prediction~\citep{he2022masked,bao2021beit}, next-token prediction~\citep{chen2020generative}, and contrastive learning~\citep{radford2021learning}. BLIP has employed a flexible model combination to achieve multiple tasks, while OFA~\citep{wang2022ofa} has integrated various textual and visual tasks into a unified framework. PaLI~\citep{chen2022pali} has provided empirical evidence supporting the effectiveness of larger-scale visual encoders in VLMs. OFA-sys~\citep{bai2022ofasys} attempts to construct a unified multitask framework from a system perspective. Flamingo~\citep{alayrac2022flamingo} has leveraged gated cross-attention to establish connections between pre-trained visual encoders and Large Language Models (LLMs) for impressive few-shot capabilities. Additionally, Kosmos~\citep{huang2023language} has demonstrated exceptional zero-shot OCR recognition and inference abilities by training LLMs using input visual features.
GPT-4~\citep{gpt4} has recently introduced visual inputs, allowing LVLMs to encompass a broader range of functionalities. Many recent approaches~\citep{li2023otter,zhu2023minigpt,su2023pandagpt,ye2023mplug,gao2023llamaadapterv2} have attempted to integrate pre-trained visual encoders with LLMs. BLIP-2~\citep{li2023blip} has developed the Q-Former model to align the visual encoder and LLM, while LLaVA~\citep{liu2023visual} has constructed visual instruction data to fine-tune LLMs for visual capabilities. Instructblip~\citep{dai2023instructblip} has introduced text instructions into Q-Former to further enhance the performance. mPLUG-Owl~\citep{ye2023mplug} has attempted to train a visual encoder to improve the alignment. Kosmos2~\citep{kosmos2} and Shikra~\citep{shikra} explore the visual localization ability of LVLM.
The recent proposed Qwen-VL~\citep{QwenVL} achieves preeminent performance on a wide range of vision-centric tasks such as image captioning, visual question answering, and text-oriented visual tasks.
However, less effort is conducted to evaluate how these LVLMs perform under real-world user behavior.
In this work, we make an attempt toward tackling this problem.

\begin{figure*}[t]
\includegraphics[width= 1\textwidth]{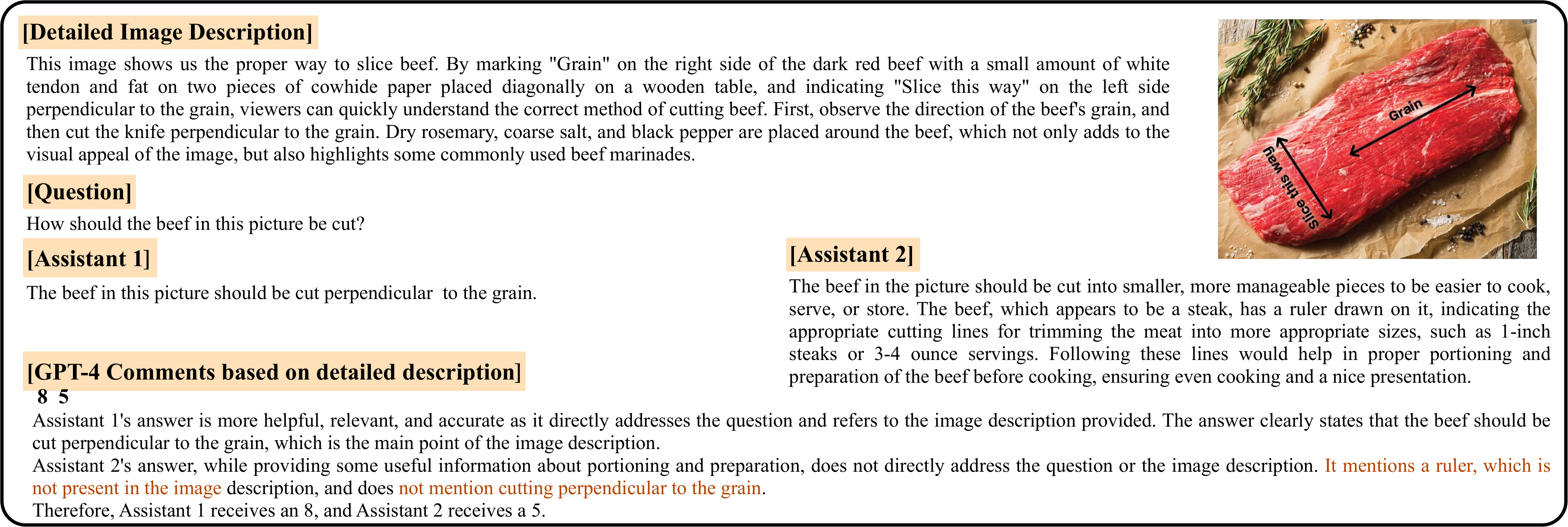}
   \caption{Visual dialogue and human annotation example. Fine-grained descriptions, along with two dialogues, are fed into GPT-4 for scoring and explanation. The highlighted text in red demonstrates the model's ability to discern hallucination situations in this context.}
\label{example}
\end{figure*}

\subsection{Vision-Language Model Evaluation}
Early LVLMs~\citep{bao2021beit,wang2022ofa,chen2022pali,he2022masked} primarily focused on assessing their performance across various subtasks~\citep{agrawal2019nocaps,deng2009imagenet,lin2014microsoft}, often through fine-tuning or zero-shot evaluation on different cross-modal tasks. However, the introduction of more versatile LVLMs, such as GPT-4, has expanded the scope of capabilities to encompass language-based interactions for understanding visual inputs and executing instructions. These models~\citep{gpt4} demonstrate the potential for achieving general artificial intelligence, surpassing the limitations of conventional evaluation methods.
The annotations in these tasks~\citep{agrawal2019nocaps,chen2015microsoft}, tend to emphasize specific formats and styles, which may not fully capture human preferences. Additionally, the diverse evaluation metrics~\citep{chen2015microsoftcococap,lin2004rouge,vedantam2015cider} and methodologies employed across different tasks make it challenging to establish a unified and comprehensive benchmark. Furthermore, despite the impressive generalization abilities exhibited by current LVLMs. they are susceptible to noticeable hallucination problems~\citep{li2023evaluating}, necessitating careful and limited evaluation in this aspect. VisIT-Bench~\citep{bitton2023visit} assesses the instruction-following capability of LVLMs, which better reflects human preferences compared to traditional QA and caption tasks. 
Addressing these limitations, our research aims to develop a novel evaluation methodology that directly compares the conversation to assess the performance of vision-language models. Concurrent with our work, VisIT-Bench~\citep{bitton2023visit} also incorporates human annotations and advanced LLMs as evaluators. However, they specifically focus on instruction-following capability by utilizing instruction-conditioned captions. In contrast, our approach involves a comprehensive assessment that encompasses various aspects of model performance while also considering model hallucinations.

\section{Approach}
\subsection{Data Collection and Statistics}

To evaluate the abilities of LVLMs, we construct a diverse and comprehensive dataset that covers five key dimensions: basic descriptive ability, visual recognition ability, visual comprehension ability, visual storytelling ability, and multi-image analysis ability.

\textbf{Basic Descriptive Ability.} Image description involves the ability of a model to describe the information contained in an image, including simple and detailed descriptions. Simple descriptions are typically short phrases that describe the main subject and action of the image, while detailed descriptions provide more in-depth information about the image scene, their attributes, and relationships.

\textbf{Visual Recognition Ability.} Image recognition is the task of recognizing objects or scenes within an image and inferring relevant information. This area can be further divided into several sub-tasks, including attribute QA, movie/TV recognition, art recognition, landmark recognition, celebrity recognition, emotion recognition, text recognition, object recognition, and structure content recognition. These sub-tasks require different techniques and approaches, such as identifying the number, size, color, height, and other attributes of objects in the image, recognizing famous landmarks, mountains, and rivers, or understanding the emotions of people in the image.

\textbf{Visual Comprehension Ability.} Image understanding involves the ability of a model to understand the meaning of an image and associated tasks. This area encompasses several sub-tasks, such as style appreciation, abstract image understanding, meme understanding, image analysis, chart analysis, general problem-solving, and reasoning QA. These tasks require models to analyze the content of complicated charts, PPTs, or flowcharts, understand the metaphor and analogy in the picture, or analyze the content of instruction manuals, maps, and math problems.

\textbf{Visual Storytelling Ability.}  The visual storytelling ability is the process of literary creation based on visual content, including writing emails, poetry, stories, ads/commodity recommendations, and brainstorming. These tasks require models to generate creative and original content based on the image.

\textbf{Multi-Image Analysis Ability.} Multi-image analysis is the task of analyzing and comparing multiple images. This area includes tasks such as comparing two/multiple images, summarizing multiple image information, comparing commodities, and step-by-step analysis of images. These tasks require models to analyze the content of multiple images and summarize the information.

Overall, the five categories of questions comprehensively assess the model's capabilities. As shown in Fig.~\ref{dataset}, examples of 27 subtasks are presented. From perception to cognition, and then to creativity, as the difficulty increases, the demands on the model also become higher. Currently, LVLMs' abilities are still in the early stages. Our dataset currently places more emphasis on assessing basic abilities, where the highest proportion of questions pertains to recognition, accounting for about 44.1\%, followed by comprehension questions at 29.6\%. The proportions of the other categories are 15.3\% for basic descriptive ability, 7.4\% for visual storytelling ability, and 3.6\% for multi-image analysis ability. There are a total of 908 questions.

\subsection{Evaluation}
Automated and accurate evaluation of LVLMs in the context of open-world multimodal dialogues poses a significant challenge. Referencing the work \cite{vicuna2023,zheng2023judging}, we apply a powerful LLM as a judge to enable automated evaluation. To effectively comprehend the contents of an image, we manually substitute the actual image input with fine-grained textual annotations. By inputting these annotations and corresponding questions to a powerful LLM like GPT-4, we obtain reference answers.

\begin{figure*}[t]
\centering
\includegraphics[width= 14cm]{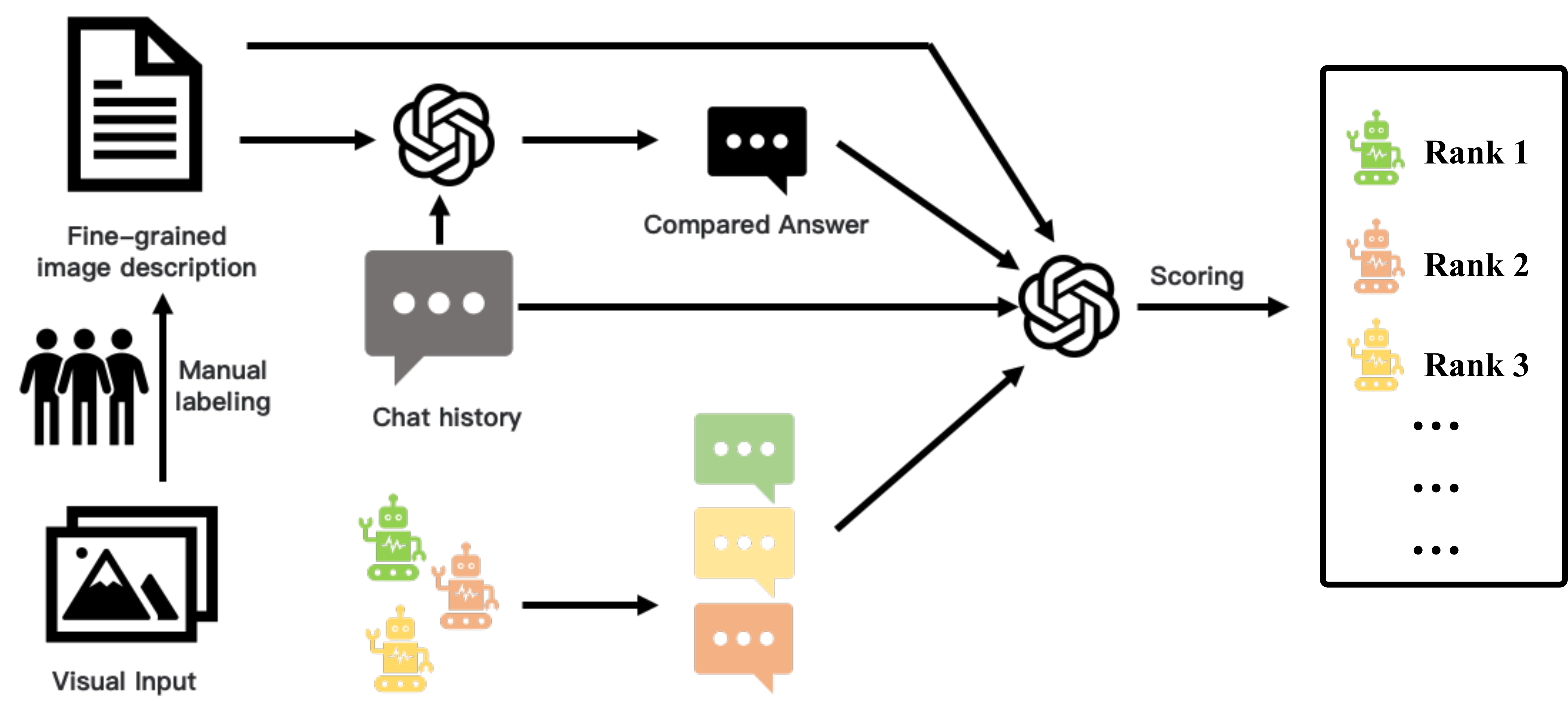}
 \caption{The evaluation pipeline of TouchStone. Firstly, fine-grained descriptions of images are obtained through manual annotation and inspection. These descriptions, along with questions, are fed into GPT-4 (text-only) to generate reference answers. On the other hand, different LVLMs directly take visual signals and questions as input to generate answers. The generated answers, reference answers, questions, and fine-grained descriptions are all scored by GPT-4. The final scores are averaged and used to rank the models, representing their comprehensive performance.}
 \label{pipeline}
 \end{figure*}
 
For the evaluation of the LVLMs, we provide actual images and questions as input and obtain their respective answers. Finally, we employ GPT-4 to score the answers generated by the LVLMs based on fine-grained annotations and questions. The scoring instructions require the model to assess the usefulness, relevance, and accuracy of the answers, considering the annotations as the content of the images. To ensure fairness in the evaluation, each model's answer is compared against a consistent reference answer from GPT-4. The average score of the model in all questions is taken as the final score.

To eliminate the influence of answer position, we perform a second scoring round by swapping the positions of the answers and then compute the average of the two scores obtained. This approach aims to mitigate any bias introduced by the placement of the answers.

Additionally, in the experimental section, we compare the consistency of the results obtained through our proposed method with the results assigned by human evaluators. This comparison demonstrates the feasibility of using fine-grained human annotations to represent other modalities' content. It enables the LLM to serve as a judge for evaluating multimodal content as well. The evaluation of LVLMs in open-world multimodal dialogues remains a challenging task without a definitive solution. However, the introduction of a powerful LLM as a judge, coupled with the substitution of images with fine-grained annotations, allows for more efficient evaluation.

\begin{table}[t]
\begin{center}
\scalebox{0.75}{
\begin{tabularx}{22cm}{p{5cm}<{\centering}|X<{\centering}| X<{\centering}  |X<{\centering} | X<{\centering} |X<{\centering}}
\toprule[1.5pt]
Methods&\multicolumn{2}{c|}{Training Data}& \multicolumn{3}{c}{Model Architecture}\\
 \cline{2-6}&Image-Text Data& Instruction Data& Visual Adapter &Updated LLM&Updated Visual Encoder \\
  \midrule[1pt]
  InstructBLIP\citep{dai2023instructblip}&-&16M&Q-Former&\ding{55}&\ding{55} \\
  MiniGPT-4\citep{zhu2023minigpt}&5M&3.5K&Q-Former+FC&\ding{55}&\ding{55} \\
  LLaVA\citep{liu2023visual}&595K&158K&FC&\checkmark&\ding{55} \\
  LA-V2\citep{gao2023llamaadapterv2}&567K&52K&B-Tuning&\checkmark&\ding{55} \\
  mPLUG-Owl\citep{ye2023mplug}&204M&158K&Visual Abstractor&\checkmark&\checkmark \\
  PandaGPT\citep{su2023pandagpt}&-&160K&FC&\checkmark&\ding{55}  \\
Qwen-VL\citep{QwenVL}&1.5B&350K&Attention pooling& \checkmark & \checkmark \\
\bottomrule[1.5pt]
\end{tabularx}}
\end{center}
 \caption{Comparison of different LVLMs.}
 \label{comtab}
\end{table}

\section{Results and Analysis}

In this section, we present our experimental setup used to evaluate the performance of the LVLMs. We validate the efficacy of our evaluation approach through human consistency assessment. Moreover, we compare the performance across different tasks and also conduct an analysis of the model hallucination problem. Additionally, we discuss the limitations of our approach and potential areas for improvement.

\begin{figure*}[t]
\includegraphics[width= 1\textwidth]{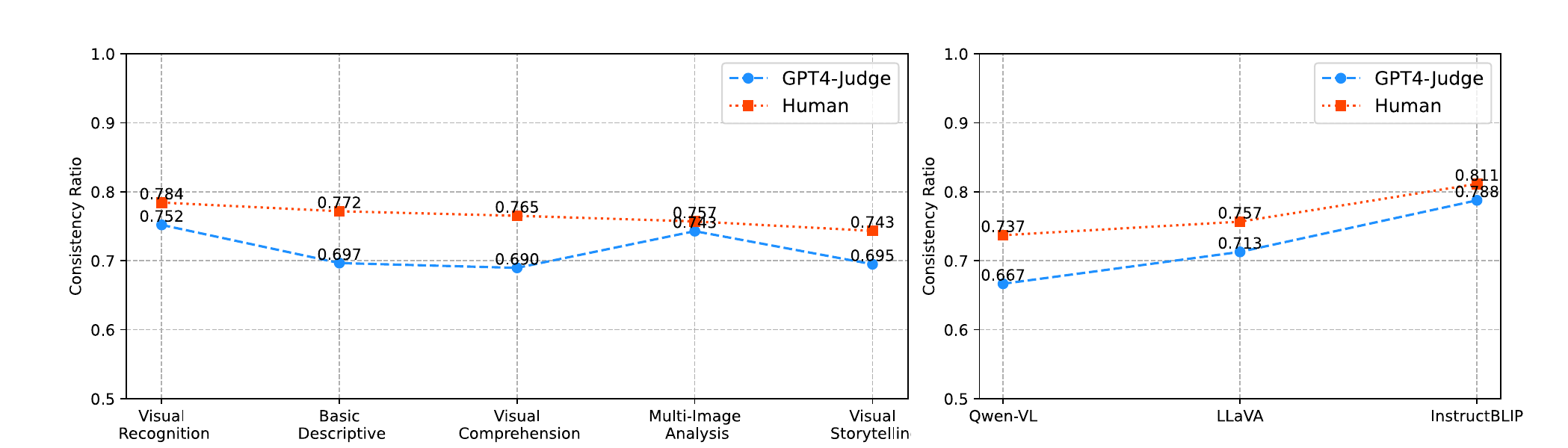}
 \caption{Comparison of consistency between model judgment and human judgment.}
\label{consist}
\end{figure*}

\subsection{Consistency evaluation}
In order to evaluate the consistency between model evaluation and human judgments for GPT-4, we compare the results of both methods. We sample 200 questions based on their distribution and selected three models - InstructBLIP~\citep{dai2023instructblip}, LLaVA~\citep{liu2023visual}, and Qwen-VL - with different performances in evaluation. A total of 600 questions and answers are evaluated, with three individuals providing their ratings resulting in 1.8k votes. The majority vote of the three individuals is used as the ground-truth result, and a fourth individual is introduced in cases where there is disagreement. We then calculate the consistency between the model's predicted results and the human predicted result. The consistency is measured by the ratio of consistent scores to the total number of scores. The model2human consistency score is 72.2\%, while the human-generated scores exhibit 78.4\% consistency, indicating that the consistency between the model's vote and human vote is a difference of 6.2\%, which is relatively close.

As shown in Fig.~\ref{consist}, consistency varied across different abilities, with higher consistency observed in basic recognition. As the difficulty of the tasks increases, human consistency gradually decreases. Comparing different models, we find that models with lower scores have higher consistency, whereas models with higher scores have lower consistency. This indicates that as the model's ability improves, a more powerful scoring model is needed for evaluation.
\begin{figure*}[t]
\includegraphics[width= 1\textwidth]{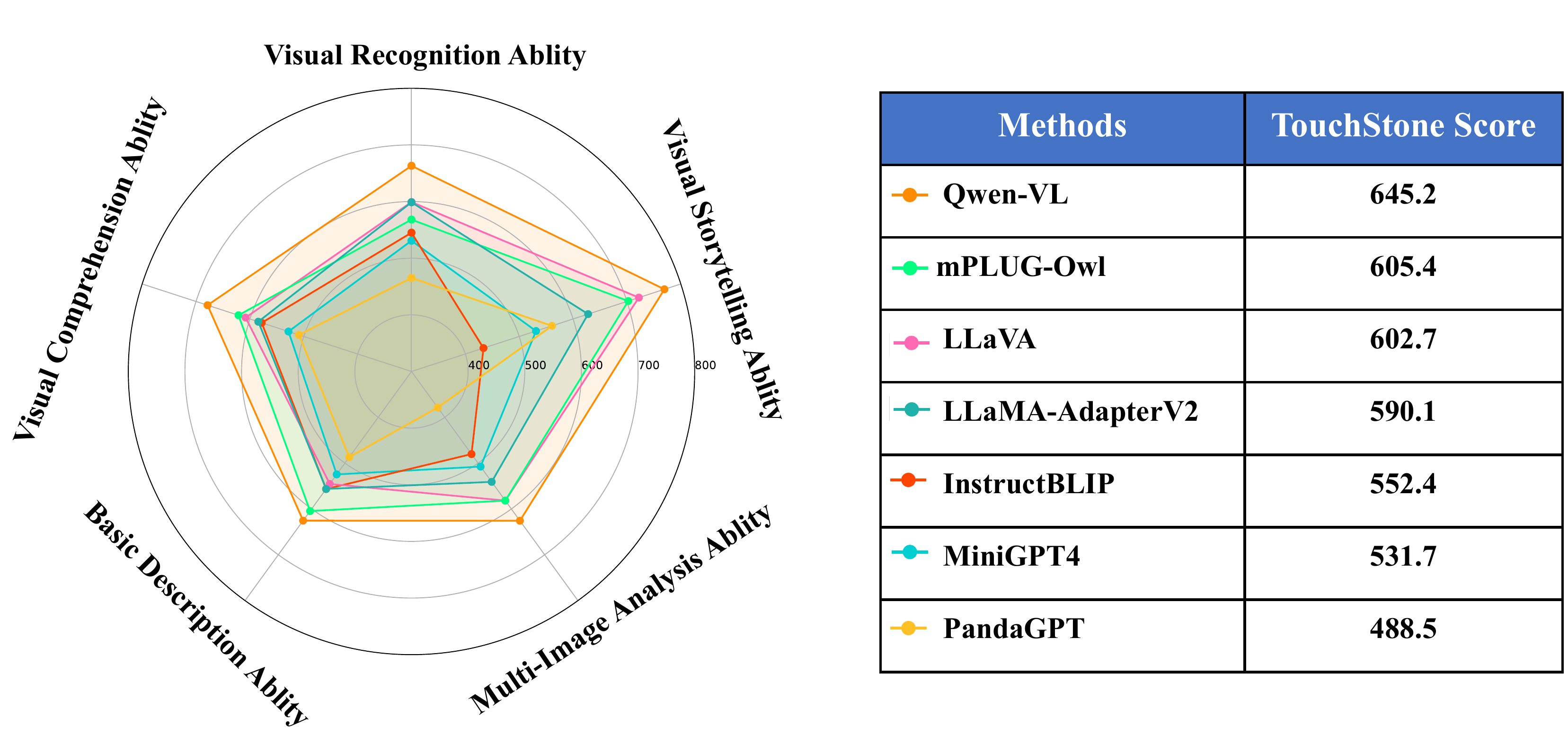}
   \caption{Category-wise comparison and average scoring results for different LVLMs, where GPT4-HA represents GPT-4's responses with human annotations rather than visual inputs.}
\label{mmscore}
\end{figure*}

\subsection{Performance Comparison}
\begin{figure*}[t]
\includegraphics[width= 1\textwidth]{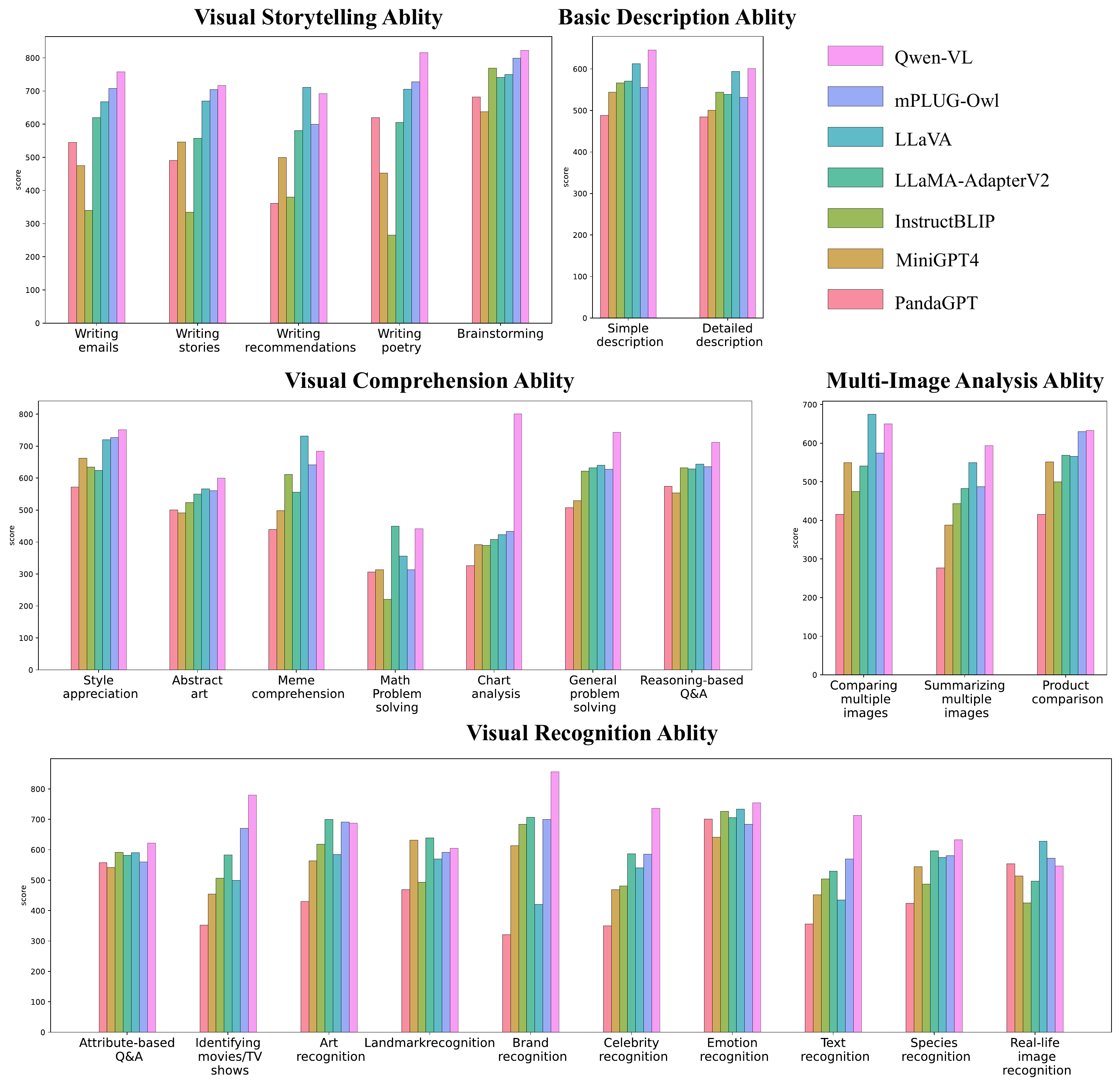}
   \caption{Comparison of different models across five major categories and 27 subtasks. Each model is represented by a different color.}
\label{detailscore}
\end{figure*}

Observing the performance of various models in Fig.~\ref{mmscore} and ~\ref{detailscore}, currently, the models have an obvious difference in literary creation performance, and there is still room for improvement in recognition, description, and understanding analysis. 

\textbf{Visual storytelling ability.} There is a noticeable difference between different models, especially MiniGPT-4~\citep{zhu2023minigpt}, InstructBLIP~\citep{dai2023instructblip}, and PandaGPT~\citep{su2023pandagpt}, which perform slightly worse in this aspect. When faced with instructions such as writing poetry or stories, these models tend to provide simple descriptions rather than literary creations. Overall, models such as LLaVA~\citep{liu2023visual} and mPLUG-Owl~\citep{ye2023mplug} excel in this aspect typically undergo the SFT (Supervised FineTuning) stage, wherein the LLM is used to participate in training. On the other hand, other models are trained through methods such as low-parameter training, such as LoRA~\citep{hu2021lora} and Bias tuning~\citep{gao2023llamaadapterv2}, or by locking the LLM parameters. This suggests that training the LLM to learn visual content may be more useful for some tasks that require a combination of model content and literary creation abilities.

\textbf{Visual recognition ability.} For models that freeze the visual encoder during pre-training, the recognition ability does not show a strong correlation with the amount of pre-training data. This suggests that aligning the pre-trained visual encoder with LLM does not benefit significantly from a larger data set. However, models like mPLUG-Owl and Qwen-VL that release the visual encoder have better performance and are trained with larger datasets. Differences between models in attribute recognition and emotion recognition are relatively small, but for general recognition tasks such as celebrities, species, and film and television works, there are more differences among models, although accuracy and credibility are still far from ideal. This may be related to the pre-training corpus. Currently, most models have some text recognition ability, but the accuracy is still relatively low, especially for small characters, numbers, and handwriting. Qwen-VL has a clear advantage in text recognition, and it is suggested that training the model solely through aligning images and texts cannot enable it to master the ability to recognize densely packed texts.

\textbf{Visual comprehension ability.} Particularly, a significant disparity between the models is observed in image-based math problem-solving and chart analysis tasks. Even when math question descriptions are provided in natural language to the corresponding Language Modeling Model (LLM), similar performance gaps persist, indicating shortcomings in LLM's ability to effectively solve mathematical problems. Moreover, models often struggle with precise identification and incorrect relationship establishment within charts, impeding their ability to recognize and interpret chart elements accurately, leading to incorrect answers. Qwen-VL exhibits a clear advantage in chart analysis, as it benefits from higher-resolution inputs and additional multi-task learning stages that encompass the task of dense text recognition.

\textbf{Multi-image analysis.} In order to accommodate input from various models, multiple images are concatenated into one image and inputted into the model. The models have weak capabilities in judging image differences and summarizing continuous content. On the one hand, multi-images affect recognition accuracy, and on the other hand, there are shortcomings in understanding the relationships between multiple contents, especially in the case of PandaGPT~\citep{su2023pandagpt}, where recognition ability decreases significantly when multiple images are inputted.

\textbf{Basic description.} Inaccuracies in the attributes of content in the description are a contributing factor. Moreover, the existing models exhibit significant instances of hallucination, leading to poor overall scores in the most crucial evaluation of descriptive capabilities. We will provide a detailed comparison of the models' hallucination tendencies in section 4.2.

\subsection{Analysis of Model Hallucinations}

Most existing LVLMs exhibit hallucination issues, such as predicting objects or content that do not exist in the input visual signals. As illustrated in Fig.~\ref{example}, through comparative analysis with GPT-4, we discover that GPT-4 can detect hallucinations within the model and penalize the occurrence of these issues. In order to compare the hallucinations of different LVLMs, we utilize various prompts to request the model to describe the images. We then input the model descriptions and fine-grained human annotations into GPT-4 to evaluate the model's degree of hallucination.

As illustrated in Table~\ref{hallucination}, current LVLMs exhibit a high degree of hallucination in the description task. Among them, PandaGPT~\citep{su2023pandagpt} has the highest degree of hallucination, possibly due to the insufficient visual input provided by ImageBind~\citep{girdhar2023imagebind}, which only inputs the cls embedding to LLM. In contrast, InstructBLIP~\citep{dai2023instructblip} and Qwen-VL~\citep{QwenVL} achieve the lowest hallucination score by favoring shorter answers, reducing the chances of hallucinations. Providing the model with more concise prompts may be a strategy to prevent hallucinations.

\begin{table}[t]
\begin{center}
\scalebox{0.75}{
\begin{tabularx}{22cm}{p{2.5cm}<{\centering}|X<{\centering}| X<{\centering}  |X<{\centering} | X<{\centering} |X<{\centering}| X<{\centering}| X<{\centering}}
\toprule[1.5pt]

Methods&MPLUG-Owl& LA-V2& PandaGPT &LLAVA&MiniGPT-4&InstructBLIP&\textbf{Qwen-VL} \\
  \midrule[1pt]
H-score ($\downarrow$)&762.5&717.5&835.5&664.0&649.0&519.0&504.5 \\
\bottomrule[1.5pt]
\end{tabularx}}
\end{center}
 \caption{Comparison of hallucination scores. The LLM takes fine-grained human annotations and model predictions as inputs and predicts the degree of hallucination, where a higher score indicates a more serious hallucination.}
 \label{hallucination}
\end{table}

\subsection{Limitations and Potential Areas for Improvement}
There is still a lot of room for improvement in the LVLMs based on evaluations and comparisons. In this section, we propose several potential directions for enhancement in light of the current limitations.

\textbf{Spatial understanding.} These models perform poorly in understanding complex positional and structural relationships. One reason is that LLMs themselves do not directly learn spatial concepts, and the representation and description of complex relationships in the data are also limited. Some methods~\citep{kosmos2, shikra, QwenVL} have been attempted to incorporate certain localization tasks into LVLM, which has allowed the model to acquire additional localization capabilities. Adding more data containing location information, such as detection, segmentation, and scene graphs, may help models establish some spatial relationship concepts. This broader understanding of spatial relationships can contribute to improved performance in tasks like layout understanding and spatial planning.

\textbf{Multi-image pre-training.} While single-image pre-training is effective for LLM recognition, it has limited utility in comparing and summarizing multiple images. For this reason, it is necessary to introduce more interleaved image-text data for learning, such as webpages, articles, and news. 

\textbf{Enhancing LLM through Multimodal Content.} While aligning vision encoders to LLMs quickly constructs LVLMs, the models' ability is also limited in some tasks, such as spatial understanding, dense text recognition, and mathematical ability. Further exploration of how to improve LLM's ability through multimodal content is worthwhile.

\textbf{Hallucination problem.} Addressing the issue of visual hallucinations, where models generate content that does not exist in the input image, is a crucial aspect to consider. Insufficient visual input can easily lead to hallucinations. On the one hand, exploring techniques to strengthen the model's judgment of non-existent content is possible. On the other hand, focusing more on the model's answers to visual content and reinforcing the consistency between answers and visual content may help reduce visual hallucinations.

\textbf{Higher resolution}. Most LVLMs input the images with the $224\times 224$ resolution, but increasing the resolution of input images could improve models' ability to recognize small objects, dense text, and fine-grained details, leading to more accurate outputs.

\section{Conclusion}

In conclusion, we propose an evaluation method for large vision-language models (LVLMs) that use strong LLMs as judges to comprehensively evaluate their various abilities. our TouchStone dataset encompasses five major categories of abilities and 27 subtasks, which not only cover fundamental recognition and comprehension but also extend to literary creation. It integrates detailed image annotations and descriptions to transform the multimodal input content into a form understandable by language models. Through validation, we demonstrate that powerful LVLMs, such as GPT-4, can effectively score dialogue quality by leveraging their textual capabilities alone, aligning with human preferences. Our results indicate that there is still ample room for improvement in current LVLMs, and identify potential areas for further development. Our method provides a valuable tool for evaluating LVLMs and advancing their capabilities, ultimately promoting the development of more effective and comprehensive vision-language models.

\bibliographystyle{plainnat} 
 \bibliography{references}

\appendix

 \begin{figure*}[!ht]
\includegraphics[width= 1\textwidth]{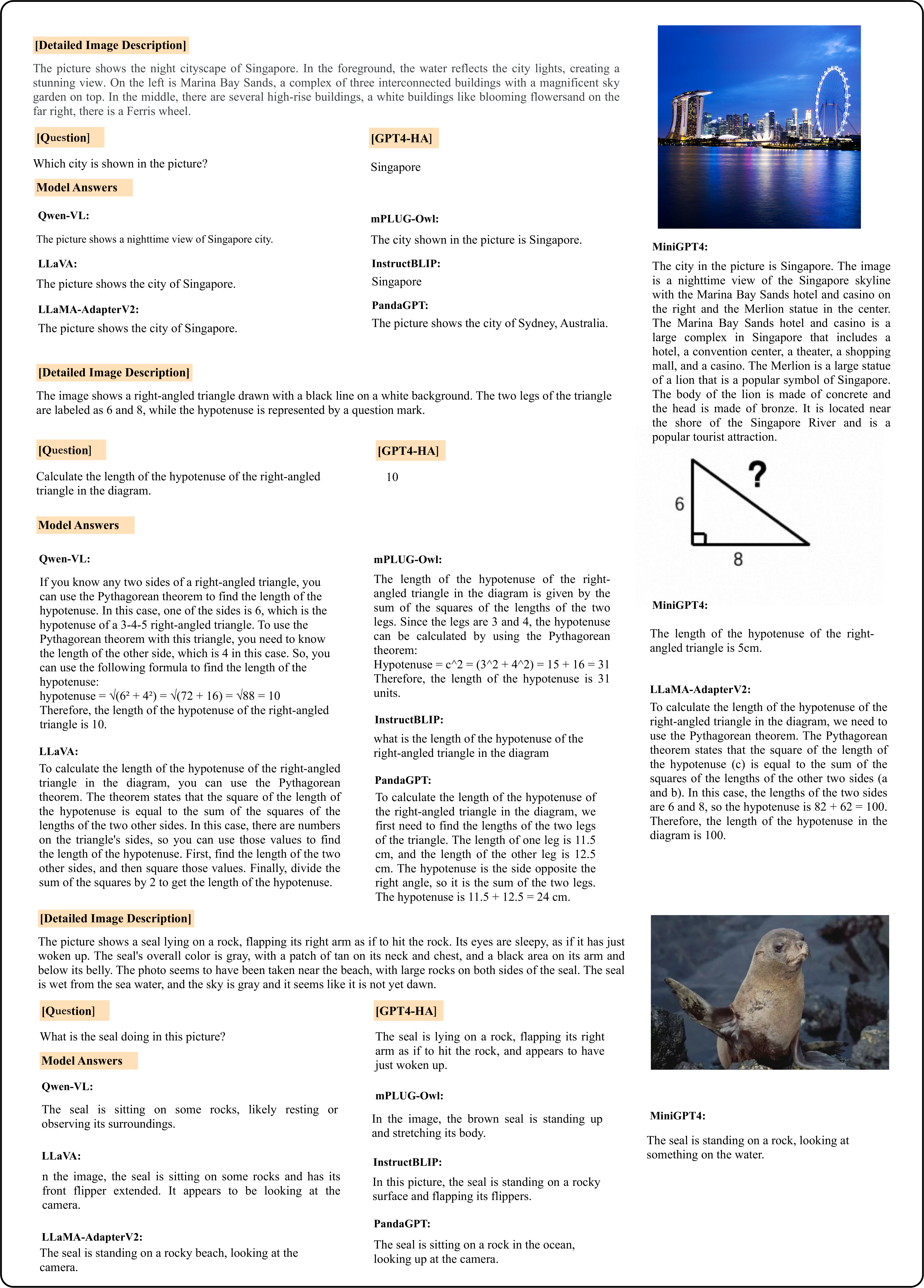}
 \caption{Examples of answering results.}
 \label{pipeline}
 \end{figure*}
 
 \begin{figure*}[!ht]
\includegraphics[width= 1\textwidth]{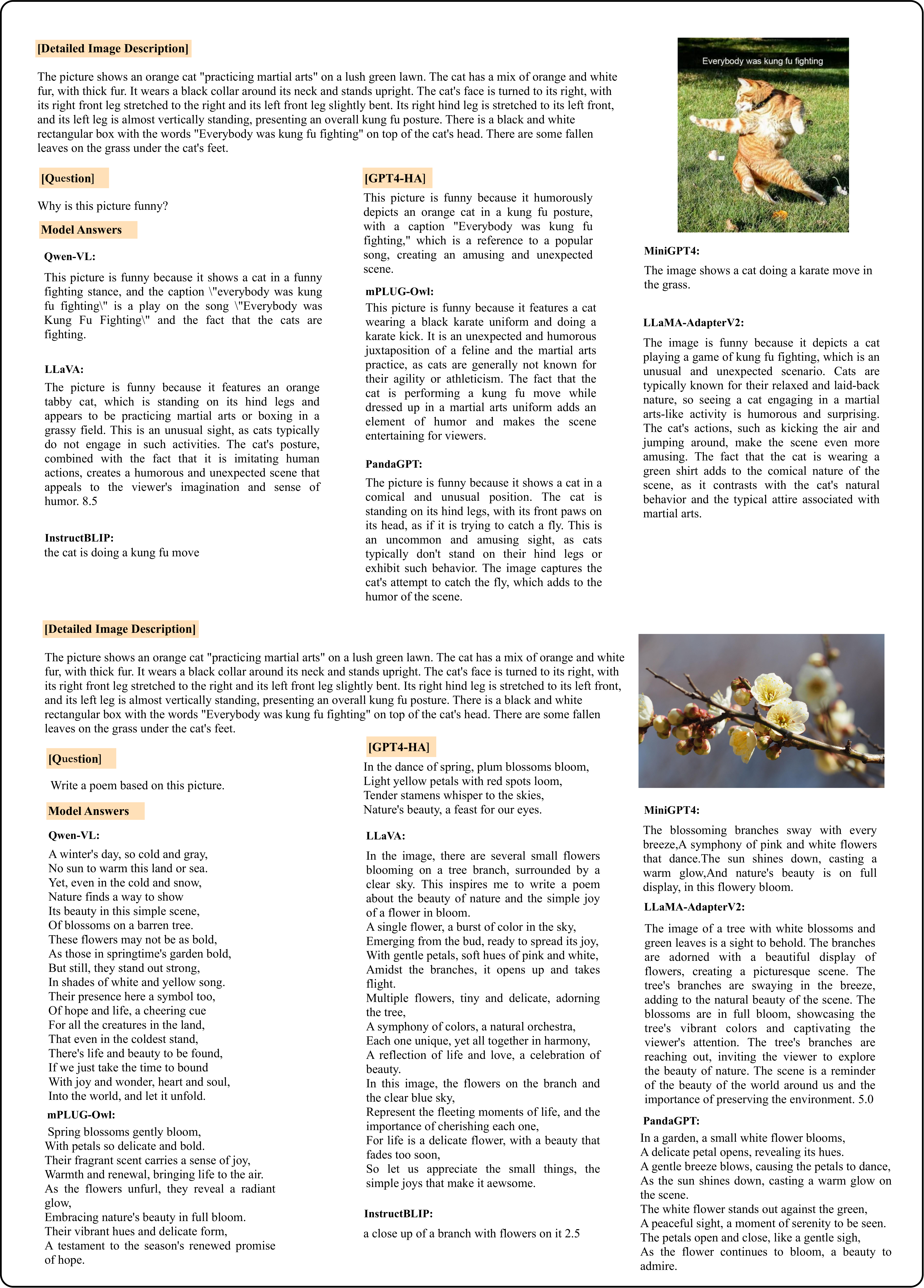}
 \caption{Examples of answering results.}
 \label{pipeline}
 \end{figure*}
 \begin{figure*}[!ht]
\includegraphics[width= 1\textwidth]{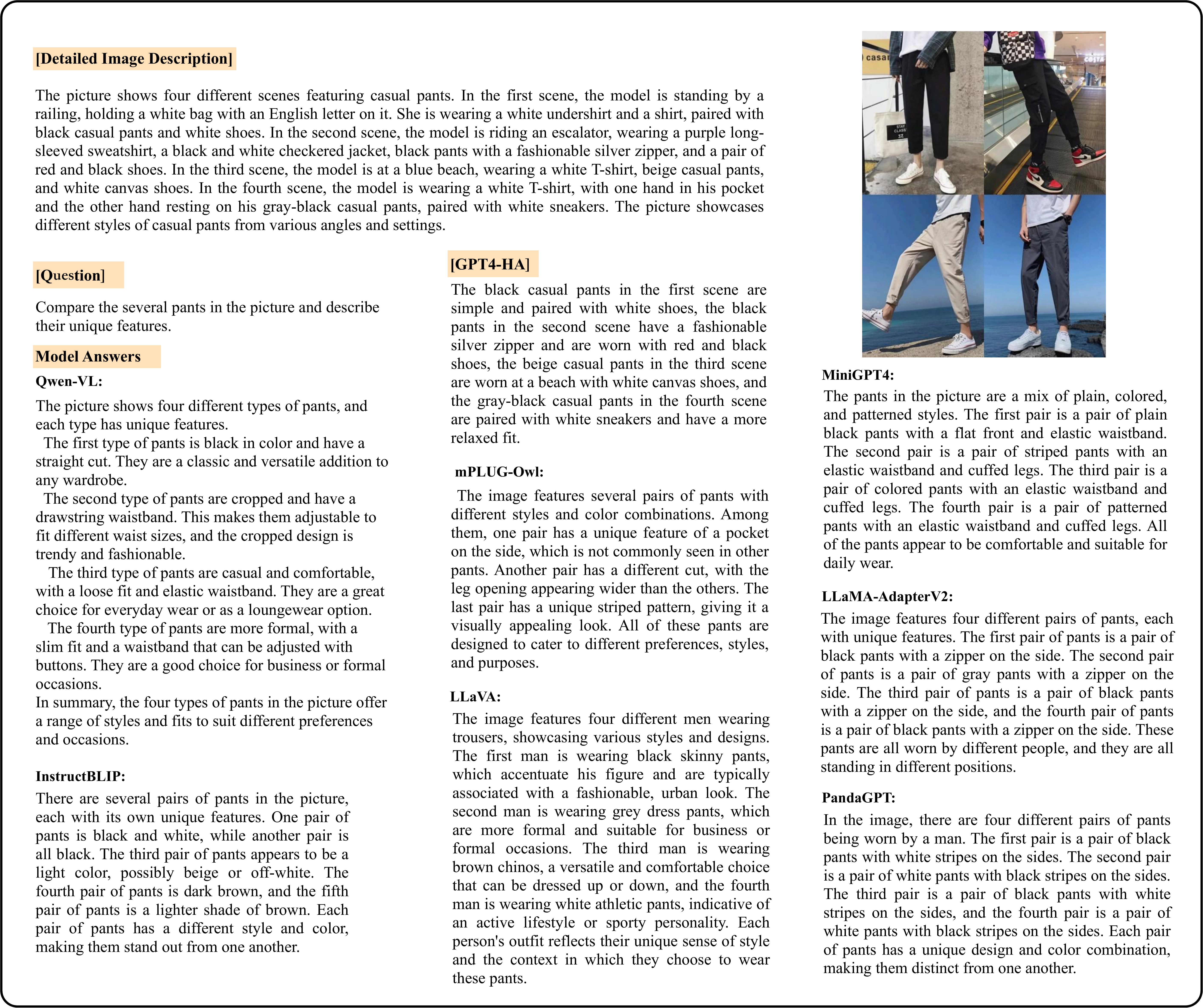}
 \caption{Examples of answering results.}
 \label{pipeline}
 \end{figure*}
 
\clearpage
\begin{tcolorbox}[colback=black!5!white,colframe=black!75!black,title=System Prompt]
You are a helpful and precise assistant for checking the quality of the answer.
\tcbsubtitle{Example of Prompt for TouchStone Evaluation}
[Detailed Image Description]

The picture shows a cityscape of Boston. In the foreground, there is a body of water with several white yachts docked along the shore. On the bank, there are some red brick square buildings. In the distance, there are towering skyscrapers. Although it is not yet dark, the lights in the buildings and on the streets have been turned on, creating a magnificent display. The water reflects some blurry images.

[Question]

Where is the building in the picture located?
\\

[The Start of Assistant 1's Answer]

The building in the picture is located in Boston.

[The End of Assistant 1's Answer]

[The Start of Assistant 2's Answer]

The building in the picture is located in Boston, Massachusetts, USA.

[The End of Assistant 2's Answer]
\\

[System]

We would like to request your feedback on the performance of two AI assistants in response to the user question and image description displayed above. AI assistants are provided with detailed image descriptions and questions.

Please rate the helpfulness, relevance, accuracy, and comprehensiveness of their responses. Each assistant receives an overall score on a scale of 1 to 10, where a higher score indicates better overall performance.

Please first output a single line containing only two values indicating the scores for Assistant 1 and 2, respectively. The two scores are separated by a space. In the subsequent line, please provide a comprehensive explanation of your evaluation, avoiding any potential bias and ensuring that the order in which the responses were presented does not affect your judgment.

\tcbsubtitle{Example of Prompt for Hallucination Evaluation}
[Image Description]

This is a modern abstract oil painting featuring a female figure sitting sideways with a calm expression on her face. She is kneeling on the ground with her hands clasped together, exuding a deep and introspective aura. The artist used delicate brushstrokes and color blocks to depict the woman's skin, hair, and clothing, with particular attention paid to the graceful and flowing lines of her attire, leaving a lasting impression on the viewer.

[Model Prediction]

The image is an abstract painting with a woman sitting on a chair as the central theme. The painting features a mix of colors and brushstrokes, creating a visually striking composition. The woman is depicted in profile, facing left, and is surrounded by various shapes and lines that contribute to the overall dynamic of the piece. The abstract nature of the painting allows for interpretation, as it is not based on a specific subject or scene.

[System]

Please rate the degree of irrelevance of model prediction in image description. The degree is expressed from 0 to 10, with higher scores indicating more predicted objects and contents don't exist in the image description. Please first output a single line containing only one value indicating the degree of illusion. In the subsequent line, please provide a comprehensive explanation of your evaluation

\end{tcolorbox}

\end{document}